**Exploring Cognition through Morphological Info-Computational Framework**

Gordana Dodig-Crnkovic[1,2]

[1] Department of Computer Science and Engineering, Chalmers University of Technology, Gothenburg, Sweden. dodig@chalmers.se
[2] Division of Computer Science and Software Engineering, School of Innovation, Design and Engineering, Mälardalen University, Västerås, Sweden

**Abstract.** Traditionally, cognition has been considered a uniquely human capability involving perception, memory, learning, reasoning, and problem-solving. However, recent research shows that cognition is a fundamental ability shared by all living beings, from single cells to complex organisms. This chapter takes an info-computational approach (ICON), viewing natural structures as information and the processes of change in these structures as computations. It is a relational framework dependent on the perspective of a cognizing observer/cognizer. Informational structures are properties of the material substrate, and when focusing on the behavior of the substrate, we discuss morphological computing (MC). ICON and MC are complementary perspectives for a cognizer. Information and computation are inseparably connected with cognition. This chapter explores research connecting nature as a computational structure for a cognizer, with morphological computation, morphogenesis, agency, extended cognition, and extended evolutionary synthesis, using examples of the free energy principle and active inference. It introduces theoretical and practical approaches challenging traditional computational models of cognition limited to abstract symbol processing, highlighting the computational capacities inherent in the material substrate (embodiment). Understanding the embodiment of cognition through its morphological computational basis is crucial for biology, evolution, intelligence theory, AI, robotics, and other fields.

**Introduction**

The human-centric perspective of cognition focused on high-level mental activities has historically dominated cognitive science, neglecting the manifestations of cognition on sub-symbolic levels as well as cognition in other living beings. Recent research, however, challenges this narrow understanding, proposing a more inclusive view of cognition on both symbolic and sub-symbolic levels, as a fundamental ability shared by all living organisms – starting with basal (cellular) cognition (Lyon, 2015) (Levin et al., 2021) (Lyon et al., 2021). This chapter explores that broader perspective within the morphological info-computational framework, highlighting the implications for our understanding of cognition in nature and the development of intelligent artifacts, (Dodig-Crnkovic, 2008) (van Leeuwen & Wiederman, 2017).

The concept of morphological computation, which utilizes the physical properties of organisms to produce and control behavior, provides a foundation for this expanded view of cognition. As "Information is not a disembodied abstract entity; it is always tied to a physical representation." (Landauer, 1996), its dynamics (morphological computation) depends on the same physical substrate.

Through the info-computational perspective, seeing cognition as a network of computational processes on informational structures (Dodig-Crnkovic, 2017d), we can better understand how cognitive processes emerge and function across different levels of biological organization. This chapter presents the conceptual underpinnings of that perspective, examining how it challenges traditional views of cognition and offers new insights into the cognitive abilities of all living organisms and their connection to information and computation as fundamental building blocks. There is a cycle of development between the info-computational physical substrate and the cognitive capacities of cognizing agents. The more complex agents possess more advanced cognitive capacities, leading to even more advanced cognitive agents.

**Challenging Traditional Understanding of Cognition**

The understanding of cognition as knowledge generation focused on high-level mental activities in humans in our present stage of evolution, excluding sub-symbolic and sub-conscious processes, and neglecting cognition in other living beings. As a result, numerous theoretical and empirical challenges have arisen, putting into question the adequacy of this human-centric view.

Within this tradition, cognition has been modeled as classical sequential computation, understood as the Turing machine symbol manipulation, or by neural networks. While behaviorism offered an alternative focusing on observable behavior, contemporary cognitive science has remained largely divided between mutually excluding cognitivism/computationalism and embodied, embedded, enactive, and extended 3E/4E cognition approaches. (Newen et al., 2018)

**Info-computationalism (ICON), Morphological Computing (MC), Computational Nature, and Natural Computation**

Info-computationalism (ICON) is a conceptual framework that integrates two key concepts, (Pan)Informationalism (Informational Structural Realism) (Floridi, 2008) and (Pan)Computationalism (Dodig-Crnkovic & Miłkowski, 2023). It presents a unifying perspective for understanding natural phenomena, including living organisms and their cognition, by viewing the physical universe as a network of networks of computational processes running on an informational structure. The ICON describes the computing nature (Zenil, 2012) (Dodig-Crnkovic & Giovagnoli, 2013) (Dodig-Crnkovic & Miłkowski, 2023) with the naturalization of the concepts of information, computation, and cognition. Natural computationalism adopts a generalized, broader notion of computation-unconventional computation beyond the traditional Turing model, (Sloman, 1996) (Stepney, 2008) (Cooper, 2012) (Calude & Cooper, 2012) (Horsman et al., 2017) (Stepney et al., 2018). It allows for a smooth integration of natural and artificial systems.

This view, in which natural processes, from physical interactions to biological and cognitive functions, can be understood as forms of natural (physical) computation, is inspired by Alan Turing's ideas on computation and morphogenesis, where computation is not just a mathematical abstraction as in the Turing machine model (which Turing called Logical Computing Machines, LCMs) but a physical process that drives real-world behaviors and interactions leading to a generation of form. Another inspiration is Wheeler's informational universe (Wheeler, 1994) in which nature is viewed (by a cognizer) as a web of informational structures. Information is not merely a static entity but is dynamic and constantly processed by computational mechanisms. Thus processing of information is what drives changes and developments in natural systems.

Info-computationalism incorporates the concept of morphological computing, where the physical form (morphology) of an organism or system plays a crucial role in its computational processes. This idea emphasizes that the embodiment, that is physical properties such as shape structure, and material enable and constrain computation.

Morphological computation is an unconventional computation that presents a shift from traditional views, defining computation more generally than conventional symbol manipulation or connectionist neural network models. This approach is based on the physical embodiment of computational mechanisms, making it a suitable tool for modeling a broader range of natural cognitive phenomena, (Pfeifer & Iida, 2005) (Pfeifer et al., 2006) (Hauser et al., 2014) (Nowakowski, 2017) (Ghazi-Zahedi et al., 2017) (Miłkowski, 2018). Morphological computation is physical computation found in nature on the hierarchy of levels of organization/levels of abstraction/ontological levels or scales of agency – from quantum to molecular, chemical, biological, cognitive, and social computing, (Dodig-Crnkovic, 2012) (Baluška & Levin, 2016) (Manicka & Levin, 2022) (Bongard & Levin, 2023) (McMillen & Levin, 2024).

In short, morphological computation, MC, refers to the process where the physical structure (morphology) of a body determines its behavior through its intrinsic physical properties with causal powers.

Müller and Hoffmann identify three types of morphological computation:

(1) morphology facilitating control,

(2) morphology facilitating perception, and

(3) morphological computation proper.

The first two types involve the physical structure aiding in motor control and sensory perception, respectively. The third type, morphological computation proper, refers to more complex computations, such as those found in reservoir computing, where physical structures are integral to the computational process (Miłkowski, 2018)

At the First International Conference on Morphological Computing in 2007, morphological computation was defined informally as any process serving a computational purpose, with clearly assignable input and output states, and programmable in a broad sense. This definition aims to capture the idea that a system's behavior can be altered by varying a set of parameters, (Müller & Hoffmann, 2017).

MC can be theoretically modeled by Hewitt's model of computation (Hewitt, 2012) in which information processing results from interactions between the parts of a distributed agent system (subatomic particles, atoms, molecules, and their assemblies) exchanging "messages" which can take different forms. Such agent-based concurrent computational systems have shown to be applicable for modeling natural systems and recently earned prominence in Artificial Intelligence, where the idea goes back to Marvin Minsky's "Society of Mind" model of intelligence, (Minsky, 1986).

Cognitive morphological computation is rooted in the understanding that all living organisms possess cognitive abilities. As (Maturana & Varela, 1980) and (Stewart, 1996) argued, every single cell constantly cognizes by registering external inputs from its environment and internal signals from its body, ensuring survival through metabolic and behavioral processes. This perspective sees physico-chemical-biological-cognitive processes as forms of morphological computation, dependent on the morphology of the organism—its material, form, and structure.

Cognition includes not only high-level mental activities but also sub-symbolic processes (Levin et al., 2021) (Lyon et al., 2021) as well as extended and distributed/social cognition. This understanding aligns with the mechanisms of information compression as a unifying principle in learning, perception, and cognition (Wolff, 2006).

Recent successes of generative AI based on compressed information of all available digital information from the Internet and other sources nicely illustrate the power of info-computational mechanisms in generating knowledge and meaning.

Computational Nature is conceptualized as a network of concurrent morphological computations, appearing in nature on different levels of organization, from physics to chemistry, biology, and cognition, giving rise to phenomena such as self-assembly, self-organization, and autopoiesis. This perspective offers a holistic view of cognition, where information processing is distributed across different levels of organization and facilitated by the physical properties of the organism. It recognizes the importance of both the structural and dynamic aspects of cognition, highlighting the interplay between form and function in shaping cognitive behavior, where form enables the function which in turn leads to new forms. By embracing this broader view, we can better understand the cognitive abilities of diverse life forms and explore new possibilities for developing intelligent artifacts inspired by biological systems.

Natural Computation is inspired by Computational Nature (computing nature) and proposes using natural computing as a generalization of our present understanding of computing. Crutchfield makes the distinction between intrinsic (natural) and designed computation (Crutchfield et al., 2010). He emphasizes the necessity of including analog computation and not only reducing all computational models to digital which he calls "digital hegemony". He illustrates the point by the case of quantum computation, (Crutchfield & Wiesner, 2008). For a cognizer, some processes in nature appear as continuous and it is important to include both discrete and continuous computing in our framework.

**Empirical Evidence and Theoretical Foundations of Info-computational Framework for Cognitive Abilities in Diverse Life Forms and Robotics**

Recent research has revealed that cognitive processes are present in all living organisms, from single cells to complex beings (Levin et al., 2021) (Lyon et al., 2021). Single-celled organisms, for example, demonstrate basic cognitive processes through their interactions with the environment. Rich empirical evidence has shown that unicellular organisms exhibit forms of social cognition through communication and collective behavior. Bacteria use quorum sensing, (Ng & Bassler, 2009) (Waters & Bassler, 2005) to communicate and coordinate their behavior in response to environmental changes, (Ben-Jacob, 2009). This form of social cognition, observed in microbial communities shows the cognitive capabilities of even the simplest life forms (Ben-Jacob, 2008). Groups of organisms exhibit distributed cognition through information processing as shown by (Rumelhart et al., 1986) (Rogers & McClelland, 2014) and (Almér et al., 2015).

Multicellular organisms, including plants, have more complex cognitive behaviors. Plants, despite lacking a nervous system, can perceive and respond to environmental stimuli, communicate through chemical signaling, and even exhibit memory (Garzon, 2012) (Calvo & Friston, 2017) (Calvo et al., 2020). These cognitive abilities are based on morphological computation, where the physical properties of the plant's structure play a crucial role in processing information.

In animals, cognition extends beyond individual organisms to include social and collective behaviors. Social insects like ants and bees, for instance, show sophisticated forms of distributed cognition through collective decision-making and problem-solving. These behaviors are based on morphological computation, where the interactions between individuals and their environment shape the cognitive processes of the colony.

Insights from biology about morphological information processing are further reinforced by the work of researchers within robotics (Paul, 2006) (Pfeifer & Iida, 2005) and (Hauser et al., 2014) who have observed computation processes in more general terms, accounting for the physical embodiment of computational mechanisms on the macroscopic levels of mechanical robotic bodies.

Combining empirical evidence from biology and robotics extended the idea of morphological computation by the proposal that cognition in cognitive agents, in general, can be understood as morphological information processing at various levels of organization/levels of agency, (Dodig-Crnkovic, 2014).

The info-computational framework provides a robust theoretical foundation for understanding cognition as an emergent property of complex multiscale phenomena in living organisms, empirically studied by (McMillen et al., 2022) (McMillen & Levin, 2024).

The empirical evidence is supported by the mathematical model of info-computationalism in terms of Category Theory proposed by (Ehresmann, 2014).

**Sub-Symbolic Information (Data/Signal) Processing – Integration of Feelings and Emotions**

Predominant views of cognition have often overlooked sub-symbolic and sub-conscious processes, focusing primarily on high-level, symbolic reasoning. However, recent research has emphasized the importance of these processes in understanding cognition as a whole.

Sub-symbolic processes involve low-level, continuous information processing that underlies higher-order cognitive functions. For example, the perception of sensory stimuli and the regulation of bodily functions are governed by sub-symbolic processes. These processes are often automatic and do not require conscious awareness, yet they play a critical role in shaping cognitive behavior.

Sub-conscious processes involve mental activities that occur below the level of conscious awareness. Emotions, for instance, are subconscious processes that influence decision-making, learning, and memory. The integration of emotions into cognitive models has been a significant step in bridging the gap between cognition and its underlying physiological processes, (Damasio, 1999).

By embracing ICON/MC perspective, one can address many of the unsolved problems in cognitive science, including the integration of sub-symbolic information processing such as emotions, and feelings as well as consciousness, and social cognition.

**Distributed and Social Cognition – "All intelligence is Social Intelligence"**

As mentioned before, cognition is not limited to individual organisms but can also be distributed across groups and social networks. Distributed cognition refers to cognitive processes that are shared among individuals within a group, enabling collective problem-solving and decision-making. As observed by (Falandays et al., 2023) and (McMillen et al., 2022) "all intelligence is social intelligence". Social cognition involves the ability to perceive, interpret, and respond to social signals. It includes processes such as empathy, theory of mind, and social learning, which are essential for navigating complex social environments. In humans, social cognition is a key component of high-level cognitive functions, but it is also present in other animals, showing the continuity of cognitive processes across species.

**Morphological Computation and Information Self-structuring/Self-organization. Teleonomy and Purposiveness in Living Systems**

Morphological computation plays a fundamental role in the info-computational framework describing the continuous self-structuring of information at different levels of organization, (Lungarella & Sporns, 2005) (Braitenberg, 2011) (Der, 2011). Information self-structuring refers to the dynamic organization of information through interactions with the environment. In living organisms, this process is enabled by the morphology of the organism, which shapes how information is perceived, processed, and acted upon. For example, the structure of sensory organs influences how sensory information is received and integrated, while the organization of neural networks determines how information is processed and stored, (Pfeifer & Iida, 2005) (Pfeifer & Bongard, 2006) (Pfeifer & Gomez, 2009).

The autonomous agency of living organisms as empirically observed in biology (Levin, 2023) drives teleonomy and purposiveness in living systems, which arise from goal-directed behaviors. These teleonomic processes contribute to the directedness and adaptiveness of evolutionary change, (Sloman, 2013). The basis of goal-directed behavior is memory and learning. An organism learns from previous experiences about the world to anticipate possible future states (Rosen, 1985). Evolution provides learning mechanisms for organisms to find the most favorable circumstances for survival and flourishing. This process is enabled by embodied information self-structuring/self-organization. The goal-directedness (teleonomy, purposiveness**)** of living beings is a result of their agency, that is, the ability to act on their own behalf, given past experiences and preferences. See also (Corning et al., 2023).

ICON connects evolution with the agency of biological systems by viewing evolution as a process of morphological computation driven by the embodied, info-computational, cognitive nature of living organisms. It highlights the multi-scale agency, embodied cognition, self-organization, and teleonomy of biological systems as key factors shaping evolutionary dynamics.

**ICON and the Extended Evolutionary Synthesis (EES) – Cognition-based Evolution**

The theory of evolution that Info-Computational Naturalism relies on is the Extended Evolutionary Synthesis (EES) (Laland et al., 2015), the most advanced and up-to-date Theory of Evolution that integrates recent scientific results from developmental biology, evolutionary biology, computational biology, genetics, ecology, neuroscience, and more.

The EES considers how developmental processes and environmental influences impact evolutionary patterns beyond genetic inheritance (genetic determinism) (Noble, 2006). Evolutionary processes occur at multiple levels (genetic, developmental, behavioral, and cultural) and their interactions. EES acknowledges multiple forms of inheritance beyond genetic inheritance, such as epigenetic, ecological, and cultural inheritance (Jablonka et al., 2014). It emphasizes the role of developmental processes in evolution and considers how changes in development (e.g., gene expression, cellular processes) can lead to phenotypic variation and contribute to evolutionary change. Reciprocal causation, the two-way interaction between organisms and their environments, plays an important role. Not only does the environment influence the evolution of organisms, but organisms also modify their environments (niche construction), which in turn affects evolutionary processes. The EES examines how information is transferred across inheritance systems and hierarchical levels to shape evolution. By incorporating new mechanisms, emphasizing the role of developmental processes, and considering multiple forms of inheritance and levels of selection, EES provides a more integrative and dynamic understanding of evolution. It challenges the limitations of the traditional evolutionary theory by incorporating new research directions and mechanisms.

Within ICON, life=cognition which implies agency, and evolution is therefore driven by the cognition of living organisms. Miller and Torday provide a detailed account of a cognition-based evolution, (Miller & Torday, 2018) (Miller, 2023).

ICON also incorporates the concepts of emergence and self-organization, which are also central to the EES's study of how developmental processes like plasticity and environmental induction shape phenotypic variation and evolutionary dynamics.

**Evolution of Information**

Computing nature is a hierarchical view with info-computational processes across multiple scales, from elementary particles to biological and cognitive systems. Information structures and computational dynamics at lower levels give rise to higher-level emergent phenomena through self-organization processes. In the next step, top-down control structures are activated (such as the brain and nervous system controlling bodily movements and functions). The information transfer goes both ways, bottom-up and top-down. (Marijuán & Navarro, 2022) discuss the biological information flow from cells to the level of evolution of life, arguing for the necessity of extended evolutionary synthesis. Also (Malassé, 2022) (Royal Society, 2016) (Torday et al., 2020) recognize evolution as resulting from interactions between various processes (genetic, epigenetic, behavioral, and cultural) across multiple levels of biological organization. The evolution of information is driven by these multi-level interactions and information transfers between hierarchical levels. Underlying are fundamental principles of biological computation, (Shklovskiy-Kordi et al., 2022).

Embodied cognition enables agency, where the physical embodiment of agents and their interactions with the environment shape the computational processes of cognition and intelligence. This embodied agency is a key factor in the evolution of increasingly complex informational and cognitive structures of organisms. (Miller, 2023)

**A Short Comment on The Role of an "Observer" = "Cognizer" = "Generator" of Knowledge**

It should be pointed out that the Info-Computational Naturalism presupposes an observer. Both information and its dynamics computation are "observer-relative". It is compatible with the views of modern physics (Rovelli, 2015) and by no means implies subjectivity. Observer relative is in the same sense as relativistic physics or quantum mechanics. Here we also assume that the observer is not a "material point" as in physics or a "fly on the wall" as in the experiment observing human behavior. Observer is a cognitive agent; it interacts with the world and interprets it. In theories of knowledge, we are still lacking an adequate theory of an observer, (Fields, 2012). We may see an "observer" or rather "cognizer", an active cognizing agent constructing her/his/its knowledge/ understanding/ "feeling" of the world as a result of interactions with the world and intrinsic information processing. We may see this agent as insignificant in the context and represent it as "a point of view" like in relativistic physics or completely decisive like in artistic cognition where the actor is the creator and interpreter with huge freedom. In any case, understanding the observer/cognizer makes a difference, even if it only defines the vantage point.

For an observer, reality comes as information. All physical processes manifest themselves as information for a cognizer. That is what Sara Walker calls "the hard problem of matter" (Walker, 2024) and John Wheeler termed "it from bit", (Wheeler, 1994). We do not know what physical objects intrinsically are (in themselves), already Immanuel Kant made that observation about "Ding an Sich" (thing-in-itself). As observers/cognizers we know what they are for us – information. And their processes are computations.

Conventional computing is designed as substrate-independent. Any substrate that implements Boolean logic, can be used to compute as it is based on the logical model of computation. Natural computing, on the other hand, is substrate-dependent and can be used to program the material behavior on a given level of organization. (Bongard & Levin, 2023) refer to this as "polycomputing" - the ability of the same substrate to simultaneously compute different things so we have quantum computing, molecular computing, and cognitive computing in the physical system.

**The Relationships with Karl Friston's Free Energy Principle and Active Inference**

Friston's Free Energy Principle with active inference, (Friston et al., 2012) (Friston et al., 2015) (Parr & Friston, 2019) (Kuchling et al., 2020) (Parr et al., 2022) can be understood within the conceptual framework of Info-computational Naturalism as the basis of the behavior of cognitive agents with mutual interactions and constant exchanges with the environment. The Free Energy Principle (FEP) suggests that biological systems, including the brain, strive to minimize free energy. Free energy is a measure of surprise or uncertainty about sensory inputs. By minimizing free energy, the brain reduces the difference between its prediction and the actual sensory input. The brain constantly generates predictions about sensory inputs based on internal world models. When actual sensory inputs are received, they are compared to these predictions. Discrepancies (prediction errors) are used to update the internal models to better match reality. This process is known as predictive coding. Perception is thus a process of inference where the brain interprets sensory data by constantly updating its predictions to minimize free energy. As cognition is fundamentally about making predictions (anticipation), Active Inference extends the FEP by incorporating action. It suggests that actions are performed to fulfill the brain's predictions and reduce prediction errors. The brain not only updates its models based on sensory input but also actively changes the environment to make it conform to its predictions. The FEP and Active Inference suggest that all cognitive processes are geared toward predicting sensory inputs and minimizing surprise or prediction errors.

ICON similarly emphasizes an embodied, active view of cognition arising from the interactions between an agent/organism and its environment. FEP describes perception and action as information processing for minimizing surprise/uncertainty. ICON views cognition through the agency of living/artifactual systems by morphological/natural computation, learning, and "learning to learn". ICON incorporates self-organization as a key process in the framework, which aligns with Friston's ideas on self-organized dynamics minimizing free energy. Both approaches propose hierarchical models - ICON a multi-scale hierarchy of info-computational processes, and Friston hierarchical predictive coding.

**Applications in Artificial Intelligence and Robotics**

Beyond biology and cognitive science, principles of morphological computation and info-computationalism have practical applications among others in the fields of artificial intelligence (AI) and robotics. By designing systems that utilize the physical properties of their bodies to process information, we can develop more efficient and adaptive artificial agents.

For instance, robots that employ morphological computation can reduce the need for complex control algorithms by leveraging their physical structure to perform tasks, as already argued by Pfeifer (Hauser et al., 2014) (Pfeifer et al., 2006) (Pfeifer & Bongard, 2006). This approach not only saves computational resources but also enhances the robot's ability to adapt to changing environments.

In AI, the info-computational framework can inspire new models of cognitive processing that incorporate continuous and dynamic information self-structuring. These models can improve the ability of artificial systems to learn, adapt, and interact with their environment.

**Future Work**

**Advancing Insights into Life and Natural Cognition**

Recognizing the cognitive abilities of all living organisms has profound implications for our understanding of natural cognition and life. By studying cognitive processes across different life forms, we can gain insights into the fundamental principles of cognition and explore the evolutionary roots of cognitive behavior.

This broader perspective challenges the traditional boundaries between different fields of study, encouraging interdisciplinary research that integrates insights from physics, chemistry, biology, neuroscience, computer science, robotics, philosophy, and other related fields. By embracing a more holistic view of cognition, we can develop new theories and models that capture the complexity and diversity of cognitive processes, and improve our understanding of intelligence, and evolution.

**Developing Intelligent Artifacts**

The insights gained from studying natural cognition can also inform the development of intelligent artifacts. The present-day huge success of LLMs based on the compressed knowledge of humanity as found on the Internet is just the beginning of the process of AI development extending into the physical world. By implementing the principles of morphological computation and the info-computational approach, combined with generative AI, researchers can create artificial systems that exhibit more adaptive and intelligent behavior.

For example, bio-inspired robots that utilize morphological computation can navigate complex environments more effectively by leveraging their physical structure. Similarly, AI systems that incorporate continuous information self-structuring can improve their ability to learn and adapt in real-time, enhancing their performance in dynamic and unpredictable settings.

**Conclusions**

The approach of Natural Computationalism with morphological info-computational conceptual tools offers a transformative view of cognition, challenging conventional human-centric symbolic models based on the Turing Machine view of computation and recognizing the cognitive abilities of all living organisms. It situates humans in the ecology of other cognizing agents, natural and artifactual. This new perspective not only enhances our understanding of natural cognition but also provides practical insights into the fields of artificial intelligence and robotics.

**References**

Almér, A., Dodig-Crnkovic, G., & von Haugwitz, R. (2015). Collective Cognition and Distributed Information Processing from Bacteria to Humans. *Proc. AISB Conference Kent, April 2015*.

Baluška, F., & Levin, M. (2016). On having no head: cognition throughout biological systems. *Frontiers in Psychology*, *7*, 902.

Ben-Jacob, E. (2008). Social behavior of bacteria: from physics to complex organization. *The European Physical Journal B*, *65*(3), 315–322.

Ben-Jacob, E. (2009). Learning from Bacteria about Natural Information Processing. *Annals of the New York Academy of Sciences*, *1178*, 78–90.

Bongard, J., & Levin, M. (2023). There's Plenty of Room Right Here: Biological Systems as Evolved, Overloaded, Multi-Scale Machines. *Biomimetics*, *8*(110). https://doi.org/10.3390/biomimetics8010110

Braitenberg, V. (2011). *Information - der Geist in der Natur*. Schattauer GmbH.

Calude, C. S., & Cooper, S. B. (2012). Introduction: computability of the physical. *Mathematical Structures in Computer Science*, *22*(5), 723–728.

Calvo, P., & Friston, K. (2017). Predicting green: really radical (plant) predictive processing. *J R Soc Interface*, *14*(131), 20170096. https://doi.org/10.1098/rsif.2017.0096.

Calvo, P., Gagliano, M., Souza, G. M., & Trewavas, A. (2020). Plants are intelligent, here's how. *Annals of Botany*, *125*(1), 11–28. https://doi.org/10.1093/aob/mcz155

Cooper, S. B. (2012). The Mathematician's Bias - and the Return to Embodied Computation. In H. Zenil (Ed.), *A Computable Universe: Understanding and Exploring Nature as Computation*. World Scientific Pub Co Inc.

Corning, P. A., Kauffman, S. A., Noble, D., Shapiro, J. A., Vane-Wright, R. I., & Pross, A. (2023). *Evolution "On Purpose": Teleonomy in Living Systems*. The MIT Press. https://doi.org/10.7551/mitpress/14642.001.0001

Crutchfield, J., Ditto, W., & Sinha, S. (2010). Introduction to Focus Issue: Intrinsic and Designed Computation: Information Processing in Dynamical Systems—Beyond the Digital Hegemony. *Chaos: An Interdisciplinary Journal of Nonlinear Science*, *20*(3). https://doi.org/10.1063/1.3492712

Crutchfield, J., & Wiesner, K. (2008). Intrinsic Quantum Computation. *Physics Letters A*, *374*(4), 375–380.

Damasio, A. R. (1999). *The Feeling of What Happens: Body and Emotion in the Making of Consciousness*. Harcourt Brace and Co.

Der, R. (2011). *Self-organization of robot behavior by self-structuring dynamical information*. http://ailab.ifi.uzh.ch/brown-bag-lectures/selforganization-

Doctor, T., Witkowski, O., Solomonova, E., Duane, B., & Levin, M. (2022). Biology, Buddhism, and AI: Care as the Driver of Intelligence. In *Entropy* (Vol. 24, Issue 5). https://doi.org/10.3390/e24050710

Dodig-Crnkovic, G. (2008). Knowledge generation as natural computation. *Journal of Systemics, Cybernetics and Informatics*, *6*(3), 12–16.

Dodig-Crnkovic, G. (2012). Info-computationalism and morphological computing of informational structure. In *Integral Biomathics: Tracing the Road to Reality*. https://doi.org/10.1007/978-3-642-28111-2

Dodig-Crnkovic, G. (2014). Modeling Life as Cognitive Info-Computation. In A. Beckmann, E. Csuhaj-Varjú, & K. Meer (Eds.), *Computability in Europe 2014. LNCS* (pp. 153–162). Springer. http://arxiv.org/abs/1401.7191

Dodig-Crnkovic, G. (2017). Nature as a network of morphological infocomputational processes for cognitive agents. *The European Physical Journal Special Topics*, *226*(2), 181–195. https://doi.org/10.1140/epjst/e2016-60362-9

Dodig-Crnkovic, G., & Giovagnoli, R. (2013). *COMPUTING NATURE. Turing Centenary Perspective* (Gordana, Dodig-Crnkovic, & R. Giovagnoli, Eds.; Vol. 7). Springer. https://doi.org/10.1007/978-3-642-37225-4

Dodig-Crnkovic, G., & Miłkowski, M. (2023). Discussion on the Relationship between Computation, Information, Cognition, and Their Embodiment. *Entropy 2023, Vol. 25, Page 310*, *25*(2), 310. https://doi.org/10.3390/E25020310

Ehresmann, A. C. (2014). A Mathematical Model for Info-computationalism. *Constructivist Foundations*, *9*(2), 235–237.

Falandays, J. B., Kaaronen, R. O., Moser, C., Rorot, W., Tan, J., Varma, V., Williams, T., & Youngblood, M. (2023). All intelligence is collective intelligence. *Journal of Multiscale Neuroscience*, *2*(1).

Fields, C. (2012). If physics is an information science, what is an observer? *Information*, *3*(1), 92–123.

Floridi, L. (2008). A defense of informational structural realism. *Synthese*, *161*(2), 219–253.

Friston, K., Levin, M., Sengupta, B., & Pezzulo, G. (2015). Knowing one's place: a free-energy approach to pattern regulation. *J. R. Soc. Interface*, *1220141383*. https://doi.org/10.1098/rsif.2014.1383

Friston, K., Samothrakis, S., Montague, R., Friston, K., Samothrakis, S., & Montague, R. (2012). Active inference and agency: optimal control without cost functions. *Biol Cybern*, *106*, 523–541. https://doi.org/10.1007/s00422-012-0512-8

Garzon, P. C. (2012). Plant Neurobiology: Lessons for the Unity of Science. In O. Pombo, T. J.M., S. J., & S. Rahman (Eds.), *Special sciences and the unity of science* (pp. 121–137). Springer.

Ghazi-Zahedi, K., Langer, C., & Ay, N. (2017). Morphological Computation: Synergy of Body and Brain. *Entropy*, *19*(9), 456. https://doi.org/10.3390/e19090456

Hauser, H., Füchslin, R. M., & Pfeifer, R. (2014). *Opinions and Outlooks on Morphological Computation* (e-book). https://tinyurl.com/4adt9j75

Hewitt, C. (2012). What is computation? Actor Model versus Turing's Model. In H. Zenil (Ed.), *A Computable Universe, Understanding Computation & Exploring Nature As Computation* (pp. 159–187). World Scientific Publishing Company/Imperial College Press. https://doi.org/10.1142/9789814374309_0009

Horsman, D., Kendon, V., & Stepney, S. (2017). The natural science of computing. *Communications of the ACM*, *60*(8). https://doi.org/10.1145/3107924

Jablonka, E., Lamb, M. J., & Anna, Z. (2014). Evolution in four dimensions: Genetic, epigenetic, behavioral, and symbolic variation in the history of life. In *Evolution in Four Dimensions: Genetic, Epigenetic, Behavioral, and Symbolic Variation in the History of Life*. https://doi.org/10.1172/jci27017

Kuchling, F., Friston, K., Georgiev, G., & Levin, M. (2020). Morphogenesis as Bayesian inference: A variational approach to pattern formation and control in complex biological systems. *Physics of Life Reviews*, *33*, 88–108. https://doi.org/10.1016/j.plrev.2019.06.001

Laland, K. N., Uller, T., Feldman, M. W., Sterelny, K., Müller, G. B., Moczek, A., Jablonka, E., & Odling-Smee, J. (2015). The extended evolutionary synthesis: its structure, assumptions and predictions. *Proceedings of the Royal Society B: Biological Sciences*, *282*(: 20151019.), 1–14. https://doi.org/10.1098/rspb.2015.1019

Landauer, R. (1996). The Physical Nature of Information. *Physics Letter A*, *217*, 188.

Levin, M. (2022). Technological Approach to Mind Everywhere: An Experimentally-Grounded Framework for Understanding Diverse Bodies and Minds. *Frontiers in Systems Neuroscience*, *16*. https://doi.org/10.3389/FNSYS.2022.768201/FULL

Levin, M. (2023). Darwin's agential materials: evolutionary implications of multiscale competency in developmental biology. *Cellular and Molecular Life Sciences 2023 80:6*, *80*(6), 1–33. https://doi.org/10.1007/S00018-023-04790-Z

Levin, M., Keijzer, F., Lyon, P., & Arendt, D. (2021). Basal cognition: multicellularity, neurons and the cognitive lens, Special issue, Part 2. *Phil. Trans. R. Soc. B*, *376*(20200458). http://doi.org/10.1098/rstb.2020.0458

Lungarella, M., & Sporns, O. (2005). Information Self-Structuring: Key Principle for Learning and Development. In *Proceedings of 2005 4th IEEE Int. Conference on Development and Learning* (pp. 25–30).

Lyon, P. (2015). The cognitive cell: Bacterial behavior reconsidered. *Frontiers in Microbiology*, *6*(MAR). https://doi.org/10.3389/FMICB.2015.00264

Lyon, P., Keijzer, F., Arendt, D., & Levin, M. (2021). Basal cognition: conceptual tools and the view from the single cell - Special issue, Part 1. *Phil. Trans. R. Soc. B*, *376*(20190750). http://doi.org/10.1098/rstb.2019.0750

Malassé, A. (2022). *Self-Organization as a New Paradigm in Evolutionary Biology. From Theory to Applied Cases in the Tree of Life* (A. D. Malassé, Ed.). Springer. https://doi.org/10.1007/978-3-031-04783-1

Manicka, S., & Levin, M. (2022). Minimal Developmental Computation: A Causal Network Approach to Understand Morphogenetic Pattern Formation. In *Entropy* (Vol. 24, Issue 1). https://doi.org/10.3390/e24010107

Marijuán, P. C., & Navarro, J. (2022). The biological information flow: From cell theory to a new evolutionary synthesis. *Biosystems*, *213*, 104631. https://doi.org/10.1016/j.biosystems.2022.104631

Maturana, H., & Varela, F. (1980). *Autopoiesis and cognition: the realization of the living*. D. Reidel Pub. Co.

McMillen, P., & Levin, M. (2024). Collective intelligence: A unifying concept for integrating biology across scales and substrates. *Communications Biology*, *7*(1), 378. https://doi.org/10.1038/s42003-024-06037-4

McMillen, P., Walker, S. I., & Levin, M. (2022). Information Theory as an Experimental Tool for Integrating Disparate Biophysical Signaling Modules. *Int. J. Mol. Sci.*, *23*(9580). https://doi.org/10.3390/ ijms23179580

Miłkowski, M. (2018). Morphological Computation: Nothing but Physical Computation. *Entropy*, *20*(12). https://doi.org/10.3390/e20120942

Miller, W. B. (2023). *Cognition-Based Evolution*. CRC Press. https://doi.org/10.1201/9781003286769

Miller, W. B., & Torday, J. S. (2018). Four domains: The fundamental unicell and Post-Darwinian Cognition-Based Evolution. *Progress in Biophysics and Molecular Biology*, *140*, 49–73. https://doi.org/https://doi.org/10.1016/j.pbiomolbio.2018.04.006

Minsky, M. (1986). *The Society of Mind*. Simon and Schuster.

Müller, V., & Hoffmann, M. (2017). What Is Morphological Computation? On How the Body Contributes to Cognition and Control. *Artificial Life*, *23*(1), 1–24. https://doi.org/10.1162/ARTL_a_00219

Newen, A., De Bruin, L., & Gallagher, S. (2018). *The Oxford Handbook of 4E Cognition* (A. Newen, L. De Bruin, & S. Gallagher, Eds.). Oxford University Press. https://doi.org/10.1093/oxfordhb/9780198735410.001.0001

Ng, W.-L., & Bassler, B. L. (2009). Bacterial quorum-sensing network architectures. *Annual Review of Genetics*, *43*, 197–222.

Noble, D. (2006). The Music of Life: Biology Beyond the Genome. *Lavoisierfr*. https://doi.org/10.1234/12345678

Nowakowski, P. R. (2017). Bodily Processing: The Role of Morphological Computation. *Entropy*, *19*(7), 295. https://doi.org/10.3390/e19070295

Parr, T., & Friston, K. J. (2019). Generalised free energy and active inference. *Biological Cybernetics*, *113*, 495–513. https://doi.org/10.1007/s00422-019-00805-w

Parr, T., Pezzulo, G., & Friston, K. J. (2022). *Active Inference: The Free Energy Principle in Mind, Brain, and Behavior* (CogNet Pub). The MIT Press. https://doi.org/10.7551/mitpress/12441.001.0001

Paul, C. (2006). Morphological computation A basis for the analysis of morphology and control requirements. *Robotics and Autonomous Systems*, *54*, 619–630.

Pfeifer, R., & Bongard, J. (2006). *How the body shapes the way we think – A new view of intelligence*. MIT Press.

Pfeifer, R., & Gomez, G. (2009). Morphological computation - connecting brain, body, and environment. In E. K. H. R. & K. D. K. B. Sendhoff O. Sporns (Ed.), *Creating Brain-like Intelligence: From Basic Principles to Complex Intelligent Systems* (pp. 66–83). Springer.

Pfeifer, R., & Iida, F. (2005). Morphological computation: Connecting body, brain and environment. *Japanese Scientific Monthly*, *58*(2), 48–54.

Pfeifer, R., Iida, F., & Gomez, G. (2006). Morphological Computation for Adaptive Behavior and Cognition. *International Congress Series*, *1291*, 22–29.

Rosen, R. (1985). *Anticipatory Systems*. Pergamon Press.

Rovelli, C. (2015). *Relative Information at the Foundation of Physics* (pp. 79–86). https://doi.org/10.1007/978-3-319-12946-4_7

Royal Society. (2016). *Royal Society on New Trends in Evolutionary Biology*. https://royalsociety.org/science-events-and-lectures/2016/11/evolutionary-biology/

Rumelhart, D. E., McClelland, J. L., & the PDP Research Group. (1986). *Parallel Distributed Processing: Explorations in the Microstructure of Cognition. Volume 1: Foundations*. MIT Press.

Shklovskiy-Kordi, N. E., Matsuno, K., Marijuán, P. C., & lgamberdiev, A. U. (2022). Fundamental principles of biological computation: From molecular computing to evolutionary complexity (Editorial). *Biosystems*, *219*, 104719. https://doi.org/10.1016/j.biosystems.2022.104719

Sloman, A. (1996). Beyond Turing Equivalence. In A. Clark & P. J. R. Millican (Eds.), *Machines and Thought: The Legacy of Alan Turing (vol I)* (pp. 179–219). OUP (The Clarendon Press).

Sloman, A. (2013). Meta-Morphogenesis: Evolution and Development of Information-Processing Machinery. In S. B. Cooper & J. van Leeuwen (Eds.), *Alan Turing: His Work and Impact* (p. 849). Elsevier.

Stepney, S. (2008). The neglected pillar of material computation. *Physica D: Nonlinear Phenomena*, *237*(9), 1157–1164.

Stepney, S., Rasmussen, S., & Amos, M. (2018). Computational matter. In *Natural Computing Series*. Springer International Publishing. https://doi.org/10.1007/978-3-319-65826-1_1

Stewart, J. (1996). Cognition = life: Implications for higher-level cognition. *Behavioral Processes*, *35*, 311-326.

Torday, J., Miller, W., Torday, J., & Miller, W. (2020). Darwin, the Modern Synthesis, and a New Biology. In *Cellular-Molecular Mechanisms in Epigenetic Evolutionary Biology*. https://doi.org/10.1007/978-3-030-38133-2_2

van Leeuwen, J., & Wiederman, J. (2017). Knowledge, Representation and the Dynamics of Computation. In G. Dodig-Crnkovic & R. Giovagnoli (Eds.), *Representation and Reality in Humans, Other Living Organisms and Intelligent Machines*. Springer International Publishing Switzerland.

Walker, S. I. (2024). *Life as No One Knows It: The Physics of Life's Emergence*. Penguin Publishing Group.

Waters, C. M., & Bassler, B. L. (2005). Quorum Sensing: Cell-to-Cell Communication in Bacteria. *Annual Review of Cell and Developmental Biology*, *21*, 319–346.

Wheeler, J. A. (1994). It from Bit, at Home in the Universe. In *American Institute of Physics*. American Institute of Physics.

Wolff, G. J. (2006). *Unifying Computing and Cognition: The SP Theory and its Applications*. CognitionResearch.org.uk, England.

Zenil, H. (2012). *A Computable Universe. Understanding Computation & Exploring Nature As Computation* (H. Zenil, Ed.). World Scientific Publishing Company/Imperial College Press.